\documentclass[ijoc,sglanonrev]{informs4}
\usepackage{eqndefns-left} 
\RequirePackage{tgtermes}
\RequirePackage{newtxtext}
\RequirePackage{newtxmath}
\RequirePackage{bm}
\RequirePackage{endnotes}

\OneAndAHalfSpacedXII 

\usepackage{algorithm}
\usepackage{algpseudocode}
\usepackage{tikz}

\usepackage{natbib}
 \bibpunct[, ]{(}{)}{,}{a}{}{,}%
 \def\bibfont{\small}%
\usepackage{float}
\usepackage{array}
\usepackage{stackrel}
\usepackage{comment}
\usepackage{setspace}
\usepackage{amsfonts}
\usepackage{indentfirst}
\usepackage{booktabs}
\usepackage{enumerate}
\usepackage[colorlinks,
linkcolor=blue,
anchorcolor=blue,
citecolor=blue]{hyperref}
\def\bm{\boldsymbol}

\newtheorem{Condition}{Condition}

\usepackage{amssymb}
\usepackage{multirow}
\usepackage{geometry}

\usepackage{xr}
\makeatletter
\newcommand*{\addFileDependency}[1]{
  \typeout{(#1)}
  \@addtofilelist{#1}
  \IfFileExists{#1}{}{\typeout{No file #1.}}
}
\makeatother

\EquationsNumberedThrough    

\TheoremsNumberedThrough     
\ECRepeatTheorems  %

\MANUSCRIPTNO{IJOC-0001-2024.00}

\begin{document}


\RUNAUTHOR{Li et al.}

\RUNTITLE{Federated Online Learning for Multisource Data}

\TITLE{Federated Online Learning for Heterogeneous Multisource Streaming Data}

\ARTICLEAUTHORS{%
\AUTHOR{Jingmao Li\footnote{Joint first author.}}
\AFF{Department of Biostatistics, Yale School of Public Health \EMAIL{jingmao.li@yale.edu}}

\AUTHOR{Yuanxing Chen$^{*}$}
\AFF{Yau Mathematical Sciences Center, Tsinghua University, \EMAIL{chenyuanxing@tsinghua.edu.cn}}

\AUTHOR{Shuangge Ma}
\AFF{Department of Biostatistics, Yale School of Public Health \EMAIL{shuangge.ma@yale.edu}}

\AUTHOR{Kuangnan Fang\footnote{Correspondence author}}
\AFF{Department of Statistics and Data Science, School of Economics, Xiamen University \EMAIL{xmufkn@xmu.edu.cn}}
} 

\ABSTRACT{
Federated learning has emerged as an essential paradigm for distributed multi-source data analysis under privacy concerns. Most existing federated learning methods focus on the ``static" datasets. However, in many real-world applications, data arrive continuously over time, forming streaming datasets. This introduces additional challenges for data storage and algorithm design, particularly under high-dimensional settings. In this paper, we propose a federated online learning (FOL) method for distributed multi-source streaming data analysis. To account for heterogeneity, a personalized model is constructed for each data source, and a novel ``subgroup" assumption is employed to capture potential similarities, thereby enhancing model performance. We adopt the penalized renewable estimation method and the efficient proximal gradient descent for model training. The proposed method aligns with both federated and online learning frameworks: raw data are not exchanged among sources, ensuring data privacy, and only summary statistics of previous data batches are required for model updates, significantly reducing storage demands. Theoretically, we establish the consistency properties for model estimation, variable selection, and subgroup structure recovery, demonstrating optimal statistical efficiency.  Simulations illustrate the effectiveness of the proposed method. Furthermore, when applied to the financial lending data and the web log data, the proposed method also exhibits advantageous prediction performance. Results of the analysis also provide some practical insights.
}%

\FUNDING{This work was supported by the National Natural Science Foundation of China [Grants 72071169, 72233002], National Social Science Foundation of China [Grant 21\&ZD146], Fundamental Research Funds for the Central Universities [Grant 20720231060], and the Shuimu Tsinghua Scholar Program of Tsinghua University.}




\maketitle
	
\section{Introduction}
\label{sec:intro}
Federated learning (FL) \citep{mcmahan_communication-efficient_2017} is a decentralized machine learning framework for analyzing distributed multisource data. In FL, individual data sources (clients) retain their local datasets, and privacy-preserving algorithms are used to collaboratively train models without sharing raw data. Based on assumptions about data distributions and model structures, FL methods can be broadly categorized into homogeneous and heterogeneous approaches. Homogeneous FL assumes a shared global model across all sources and aggregates all local information for model training. In contrast, heterogeneous FL allows for personalized models to accommodate source-specific heterogeneity. In this study, we focus on heterogeneous FL, which is often better suited to real-world applications. Substantial efforts have been made in FL; we refer to the comprehensive reviews for more details \citep{wen_survey_2023,pei_review_2024}.

\begin{figure}
    \FIGURE
    {\includegraphics[width=0.99\linewidth]{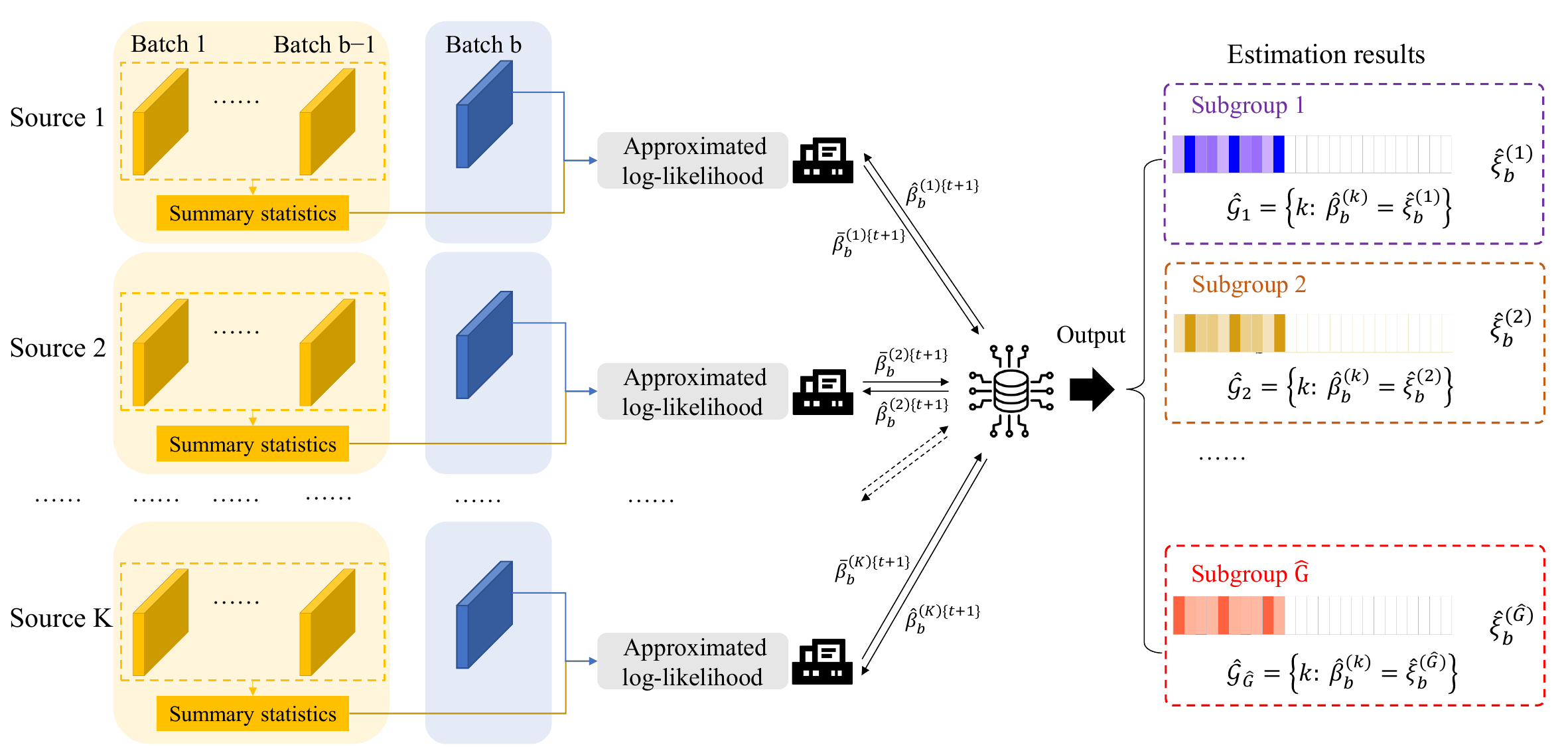}}
    {Diagram of distributed multisource streaming data and flowchart of the proposed federated online learning method. 
    \label{fig:framework}}
    {}
\end{figure}

Despite the tremendous progress, most of the existing FL methods primarily focus on ``static" data analysis. For this, data sources is capable to gather all the samples at once. However, advancements in data collection techniques have led to the emergence of multisource ``streaming data", as illustrated in left half of Figure \ref{fig:framework}. Here, data is collected from multiple sources, and each source can continuously receive new data batches (new samples) over time, significantly different from the ``static" datasets. Our study is motivated by data from a national online financial lending platform in China. This platform collects daily updated business data from various provinces (data sources) to analyze user behavior in real-time. Another example is the continuously generated web log data across multiple websites, essential for cyberattack detection. Additional potential applications include fraud detection, the Internet of Things (IoT), healthcare, and e-Commerce \citep{gomes_machine_2019}. 

Multisource streaming data presents unique challenges compared to other data types. First, real-time data collection leads to high velocity and volume, requiring efficient analysis methods that avoid excessive storage and computational costs
. Several online learning methods have been developed to analyze streaming data \citep{toulis_asymptotic_2017, luo_renewable_2020}. These methods can train and update models without accessing data from previous batches, thereby substantially reducing storage and computation demands. However, most existing approaches focus primarily on single-source data, leaving methodological and theoretical development for multisource scenarios largely unexplored. Second, as a common challenge in multisource data analysis and heterogeneous federated learning (FL), data sources often exhibit both shared and source-specific structures. Therefore, it is crucial to develop methods that can effectively balance homogeneity and heterogeneity across sources.

Motivated by the above investigations, we propose a federated online learning (FOL) estimator for analyzing high-dimensional multi-source streaming data. We consider high-dimensional settings in which only a small subset of covariates is associated with the response, necessitating variable selection. To model heterogeneous multi-source data, we assume the existence of a latent subgroup structure across sources, where sources within the same subgroup share a common subgroup-specific model. This structure facilitates the modeling of both homogeneity and heterogeneity among sources, thereby enhancing overall performance. To estimate the model, we develop an efficient algorithm based on the renewable estimation framework \citep{luo_renewable_2020} and proximal gradient descent algorithm \citep{beck_fast_2009}. The proposed method updates the model using current data and historical summary statistics, and it does not require transmission of raw data across sources. Theoretically, we establish consistency in model estimation, variable selection, and subgroup structure recovery, achieving the same convergence rate as the ``oracle” estimator and demonstrating optimal statistical efficiency. In summary, compared to existing studies, we propose a novel federated learning method for analyzing multisource streaming data, a setting that has received limited attention. Furthermore, our theoretical developments offer broader insights into the fields of federated and online learning, particularly in high-dimensional settings.


\section{Literature Review}  
\subsection{Federated Learning and multisource data analysis}
Multi-source data analysis has gained considerable attention in diverse fields, including electronic health records (EHR) modeling and customer transaction records modeling \citep{qiu_fraud_2024}. Considering the privacy, FL learning \citep{yan_federated_2024,qi_privacy_2025} has emerged as an essential paradigm for this problem. FedAvg \citep{mcmahan_communication-efficient_2017} is the most basic FL method, which trains a global model by averaging locally updated models from multiple sources after multiple rounds of iterations. Although direct, FedAvg adopts a global model for all sources, neglecting the differences among sources. In contrast, heterogeneous or personalized FL methods aim to consider model heterogeneity under FL framework. They leverage shared information across sources to build similar personalized models for related sources, thereby enhancing the performance of customized models. For example, clustered FL \citep{chen_heterogeneity-aware_2024} groups sources with similar data distributions or model parameters into clusters and trains separate models per cluster to better capture heterogeneity and improve overall performance. Knowledge distillation--based FL \citep{wang_towards_2024} enables collaboration and alignment among heterogeneous sources by having them share soft predictions or representations. Other related methods include meta learning \citep{fallah_personalized_2020} and regularization-based methods \citep{t_dinh_personalized_2020}. In addition to FL methods, other avenues of research have also investigated the multisource data analysis. For example, integrative analysis approaches \citep{huang_promoting_2017,maity_meta-analysis_2022-1} have attracted substantial interest, especially in high-dimensional scenarios. 
However, most of these methods do not explicitly address the challenges of streaming data analysis.

\subsection{Online Learning}
Traditional statistical methods typically fit models using raw data from both historical and newly arrived batches, which becomes impractical due to rapid data growth and storage constraints. In contrast, online learning enables model updates based on current individual-level data and summary statistics from previous batches, significantly reducing storage needs and the processing of extensive historical datasets. 
Among existing online learning methods, stochastic gradient descent (SGD) and its variants are well-regarded for their simplicity and effectiveness \citep{toulis_asymptotic_2017}. Despite its computational convenience, reliance on first-order gradient information may limit efficiency and statistical inference \citep{ma_general_2022}. 
Recently, \cite{luo_renewable_2020} introduced a renewable estimation method based on second-order Taylor's expansion, which has been extended to high-dimensional analysis \citep{luo_online_2023}, quantile regression \citep{xie_statistical_2024}, and others. 

A few recent studies have also focused on multisource streaming data analysis with online updates for global or source-specific model parameters. For example, \cite{hector_parallel-and-stream_2023} proposed the Parallel-and-Stream Accelerator (PASA), which uses a divide-and-combine method for distributed online learning. However, its neglect of heterogeneity and high-dimensionality limits real-world applicability. Similarly, \cite{luo_online_2022} analyzes distributed streaming data using linear mixed-effects models to account for site-specific random effects, but this method is limited to linear models and panel data, and struggles with high-dimensionality. Moreover, site-specific effects can introduce many redundant parameters, negatively impacting the estimation and prediction performance. Overall, existing methods are still insufficient in modeling high-dimensional multisource data.

\section{Methodology}
\label{sec:methodology}
\subsection{Data and Model} \label{sec:method_on}

In the context of distributed multi-source streaming data, we assume the existence of $K$ distinct sources, each continuously receiving $b$ batches of data. Denote the observations collected in the $u$-th batch from the $K$ distinct sources as $\mathcal{D}_{u}^{(k)}=\big\{ y_{u,i}^{(k)},\mathbf x_{u,i}^{(k)}\big\}_{i=1}^{n_{u,k}}$. Here, we denote $n_{u,k}$ as the sample size, $y_{u,i}^{(k)}\in\mathbb R$ as the response variable, and $\mathbf{x}_{u,i}^{(k)}\in \mathbb{R}^{p}$ as the $p$-dimensional covariate. For simplicity, we also denote $\widetilde{\mathcal D}_b^{(k)}=\{\mathcal{D}_{1}^{(k)},\dots,\mathcal{D}_{b}^{(k)}\}$ as the cumulative data for the $k$-th source up to batch $b$. 
We assume that $\mathcal{D}_{u}^{(k)}$ are independent and identically distributed across $b$ batches and $K$ sources.
Define ${N}_{b,k}=\sum_{u=1}^{b}n_{u,k}$ as the cumulative sample size for the source $k$ up to the $b$ batches, and ${N}_{b}=\sum_{k=1}^{K} N_{b,k}$ as the total sample size for all $K$ sources. 
For the $k$-th source, we assume that the conditional distribution of $y_{u,i}^{(k)}$ given $\mathbf x_{u,i}^{(k)}$ belongs to the canonical exponential family, of which the density function depends on the source-specific parameters $\bm\beta^{(k)}=(\beta_1^{(k)},\dots,\beta_p^{(k)})^\top$ and is given by:
\begin{equation}
    \label{eq:exponentail}
    f\big(y_{u,i}^{(k)}|\mathbf{x}_{u,i}^{(k)};\boldsymbol{\beta}^{(k)}\big)=c\big(y_{u,i}^{(k)};\phi^{(k)}\big)\exp\left[y_{u,i}^{(k)}\mathbf{x}_{u,i}^{(k)\top}\boldsymbol{\beta}^{(k)}-d\big(\mathbf{x}_{u,i}^{(k)\top}\boldsymbol{\beta}^{(k)}\big)\right],\quad k\in [K], u\in [b],
\end{equation}
where $d(\cdot)$ is the unit deviance function of the exponential family, $\phi^{(k)}$ is the dispersion parameter, and $c(\cdot)$ is the normalization term. For any integer $b$, we denote $[b]=\{1,\dots,b\}$. 

Given the observed data, the model parameters in \eqref{eq:exponentail} can be naively estimated by minimizing the negative likelihood:
\begin{equation}
      \label{eq:likelihood}
      \widehat{\boldsymbol{\beta}}^{(k)}_{b,\text{naive}}=\arg\min_{\boldsymbol{\beta}^{(k)}}\left\{ \mathcal L\big(\boldsymbol{\beta}^{(k)}; \widetilde{\mathcal{D}}^{(k)}_{b}\big)=\sum_{u=1}^b\sum_{i=1}^{n_{u,k}} \left[y_{u,i}^{(k)}\mathbf{x}_{u,i}^{(k)\top}\boldsymbol{\beta}^{(k)}-d\big(\mathbf{x}_{u,i}^{(k)\top}\boldsymbol{\beta}^{(k)}\big) \right]\right\}, \quad k\in [K].
\end{equation}
However, this naive estimator has several limitations. First, optimizing \eqref{eq:likelihood} requires access to all raw data up to the $b$-th batch, which, given the high velocity and volume of streaming data, imposes significant storage and computational burdens. Second, the estimator in \eqref{eq:likelihood} treats each source independently, estimating model coefficients separately and thereby ignoring potential similarities and inter-source connections. In practice, however, sources often share some homogeneous information, which can be leveraged to improve model performance -- particularly in high-dimensional settings \citep{huang_promoting_2017}.

To enhance estimation, we propose a novel Federated Online Learning (FOL) method for analyzing multi-source streaming data. Our aim is to develop a unified framework that accommodates both homogeneity and heterogeneity across data sources while adhering to federated and online learning paradigms. Specifically, following \cite{luo_renewable_2020}, we first develop a strategy to approximate the log-likelihood of previous batches ($u\in [b-1]$) for source $k$ ($k\in [K]$) using summary statistics. We first consider the case where $b=2$, in which $\mathcal D_2^{(k)}$ arrives after $\mathcal D_1^{(k)}$ for all $k\in [K]$. To avoid using individual-level data from $\mathcal D_1^{(k)}$, we can adopt following least-square approximation:
\begin{equation}\label{eq:approx1}
\begin{aligned}
\mathcal L\big(\boldsymbol{\beta}^{(k)}; \widetilde{\mathcal D}_2^{(k)}\big)=&\ \mathcal L\big(\boldsymbol{\beta}^{(k)}; {\mathcal D}_1^{(k)}\big)+ \mathcal L\big(\boldsymbol{\beta}^{(k)}; {\mathcal D}_2^{(k)}\big) \\
=& \mathcal L\big(\widehat{\bm{\beta}}_1^{(k)}; {\mathcal D}_1^{(k)}\big)+\mathbf U\big(\widehat{\bm\beta}_1^{(k)};{\mathcal D}_1^{(k)}\big)^\top\big(\boldsymbol{\beta}^{(k)}-\widehat{\boldsymbol{\beta}}^{(k)}_1\big)\\
&+\frac{1}{2}\big(\boldsymbol{\beta}^{(k)}-\widehat{\boldsymbol{\beta}}^{(k)}_1\big)^{\top} \mathbf{J}\big(\widehat{\boldsymbol{\beta}}^{(k)}_1; \mathcal{D}_1^{(k)}\big) \big(\boldsymbol{\beta}^{(k)}-\widehat{\boldsymbol{\beta}}^{(k)}_1\big)\\
& +\mathcal L\big(\boldsymbol{\beta}^{(k)}; {\mathcal D}_2^{(k)}\big)+n_1^{(k)}O_p\Big(\Big\|\widehat{\boldsymbol{\beta}}^{(k)}_1-\bm\beta^{(k)}\Big\|_2\Big),
\end{aligned}
\end{equation}
where $\widehat{\bm\beta}_1^{(k)}$ is the initial estimator from the first batch, while 
$\mathbf{U}(\boldsymbol{\beta}^{(k)}; \mathcal{D}_u^{(k)})={\partial  \mathcal L(\boldsymbol{\beta}^{(k)}; \mathcal{D}_u^{(k)})}/{\partial \boldsymbol{\beta}^{(k)}}$ and $\mathbf{J}(\boldsymbol{\beta}^{(k)};\mathcal{D}_u^{(k)})={\partial^2 \mathcal L(\boldsymbol{\beta}^{(k)};\mathcal{D}_u^{(k)})}/{\partial \boldsymbol{\beta}^{(k)}\partial \boldsymbol{\beta}^{(k)\top}}$ represent the gradient vector and Hessian matrix, respectively. 
The initial estimator $\widehat{\bm{\beta}}_1^{(k)}$ ensures $\mathbf U\big(\widehat{\bm\beta}_1^{(k)};{\mathcal D}_1^{(k)}\big)\approx \bm 0$, which allows us to asymptotically ignore the error term. Consequently, the log-likelihood function $ \mathcal{L}\big(\boldsymbol{\beta}^{(k)}; \widetilde{\mathcal D}_2^{(k)}\big)$ can be approximated as follows,
\begin{equation}\label{eq:approx2}
\widehat{\mathcal L}_2^{(k)}\big(\boldsymbol{\beta}^{(k)}\big):=\frac{1}{2}\big(\boldsymbol{\beta}^{(k)}-\widehat{\boldsymbol{\beta}}^{(k)}_1\big)^{\top} \mathbf{J}\big(\widehat{\boldsymbol{\beta}}^{(k)}_1; \mathcal{D}_1^{(k)}\big) \big(\boldsymbol{\beta}^{(k)}-\widehat{\boldsymbol{\beta}}^{(k)}_1\big)+ \mathcal L\big(\boldsymbol{\beta}^{(k)}; {\mathcal D}_2^{(k)}\big)+C,
\end{equation}
where $C$ denotes some irrelevant terms. 

Now we generalize this approximation technique to the case where $b\geq 2$. After removing the irrelevant constant, the approximated log-likelihood function
$\widehat{\mathcal L}_b^{(k)}\big(\boldsymbol{\beta}^{(k)}\big)$ can be obtained as follows,
\begin{equation}
    \label{eq:approximation}
    \widehat{\mathcal L}_b^{(k)}\big(\boldsymbol{\beta}^{(k)}\big):=\frac{1}{2}\big(\boldsymbol{\beta}^{(k)}-\widehat{\boldsymbol{\beta}}^{(k)}_{b-1}\big)^{\top} \widehat{\mathbf{J}}_{b-1}^{(k)}\big(\boldsymbol{\beta}^{(k)}-\widehat{\boldsymbol{\beta}}^{(k)}_{b-1}\big)+ \mathcal L\big(\boldsymbol{\beta}^{(k)}; {\mathcal D}_b^{(k)}\big),
\end{equation}
where $\widehat{\boldsymbol{\beta}}^{(k)}_{u}$ ($k\in [K]$, $u\in [b]$) denotes the estimate from the $u$-th batch. For $b\geq 2$, we define the matrix $\widehat{\mathbf{J}}_{b-1}^{(k)}=\sum_{u=1}^{b-1} \mathbf{J}\big(\widehat{\boldsymbol{\beta}}^{(k)}_{u}; \mathcal{D}_u^{(k)}\big)$, with $\widehat{\mathbf{J}}_{0}^{(k)}=\mathbf{0}_{p\times p}$ for consistency. 

Meanwhile, to promote and utilize the potential similarity across different sources, we adopt a novel ``subgroup" assumption among sources for analyzing multisource data. It hypothesis that there exists a subgroup partition $\mathcal G=\{\mathcal G^{(1)},\dots,\mathcal G^{(G)}\}$ such that $K$ sources can be categorized into $G$ non-overlapping subgroups, where $G\ge 2$, $\mathcal G^{(g)}\subseteq[K]$, and $\cup_{g=1}^G\mathcal G^{(g)}=[K]$, and the sources within each subgroup share common parameters. Therefore, by accurately identifying the subgroup structure, we can facilitate the information sharing within the subgroup, thereby promoting the model estimation. 

Therefore, given the approximated log-likelihood function \eqref{eq:approximation}, our FOL estimator $\widehat{\mathbf{B}}_{b}$ is obtained by minimizing the following objective function:
\begin{equation}
    \label{eq:obj} 
    \begin{aligned}
        \mathcal{Q}_{b}^{\text{on}}(\mathbf{B})=-\frac{1}{{N}_{b}}\sum_{k=1}^{K} 
        \widehat{\mathcal L}_{b}^{(k)}\big(\boldsymbol{\beta}^{(k)}\big)
        +\sum_{k=1}^{K}\sum_{j=1}^{p} \rho_{a}\left(\left|\beta_{j}^{(k)}\right|,\lambda_{1,b}\right) + \sum_{k_1< k_2} \rho_{a}\left(\left\|\boldsymbol{\beta}^{(k_1)}-\boldsymbol{\beta}^{(k_2)}\right\|_2,\lambda_{2,b}\right).
    \end{aligned}
\end{equation}
where $\mathbf{B}=(\boldsymbol{\beta}^{(1)},\dots, \boldsymbol{\beta}^{(K)})$ is a $p\times K$ coefficient matrix, $\|\cdot\|_2$ denotes the $L_2$-norm, and $\rho_{a}(x, \lambda)$ is the penalty function characterized by tuning parameters $\lambda$ and $a$.
In this paper, we adopt the minimax concave penalty (MCP) \citep{zhang_nearly_2010}, defined as  $\rho_{a}(x, \lambda)=\int_{0}^{|x|}\left(\lambda-t/a\right)_+\mathrm{d}t$ for $x\in \mathbb{R}$. The second term of $\mathcal{Q}^{\text{on}}_{b}(\mathbf B)$ in \eqref{eq:obj} serves as a sparsity penalty to select important variables, while the third term penalizes the pairwise differences between $\boldsymbol{\beta}^{(k_1)}$ and $\boldsymbol{\beta}^{(k_2)}$ (for $k_1<k_2$), therefore promoting the identification of subgroup structure. Specifically, if $\|\boldsymbol{\beta}^{(k_1)}-\boldsymbol{\beta}^{(k_2)}\|_2$ approaches $0$, then sources $k_1$ and $k_2$ are regarded to belong to the same subgroup. This fusion penalty has shown its unique advantages and is commonly employed for clustering or identification of the coefficient structure \citep{ma_concave_2017}. 

It is noted that for the objective function $\mathcal{Q}_{b}^{\text{on}}(\mathbf B)$ in (\ref{eq:obj}), we do not need to access historical data batches. Instead, we can sequentially use and update the summary statistics $\{\widehat{\mathbf{J}}_{b-1}^{(k)}, \widehat{\boldsymbol{\beta}}^{(k)}_{b-1}\}_{k\in [K]}$. Meanwhile, as discussed in subsequent sections, we optimize the objective function using federated learning, which does not require the transmission of raw data across sources. The right panel of Figure \ref{fig:framework} presents the flowchart of the proposed FOL method. Overall, the method automatically identifies subgroup structures across data sources, effectively capturing both homogeneity and heterogeneity, thereby improving estimation and prediction performance.

\subsection{Computational Algorithm} \label{sec:computation}

We implement the proximal gradient algorithm \citep{beck_fast_2009} to solve (\ref{eq:obj}). This algorithm is a variant of traditional gradient descent, capable of optimizing non-differentiable objective functions. The algorithm iterates a local calculation step under each data source and a master aggregation step under a central server, which accommodates the federated learning paradigm. Specifically, with estimated coefficients $\widehat{\mathbf{B}}^{\{t\}}_{b}=\big(\widehat{\boldsymbol{\beta}}^{(1)\{t\}}_{b},\dots, \widehat{\boldsymbol{\beta}}^{(K)\{t\}}_{b}\big)$ in the $t$-th iteration, we process the two steps as follows:
\begin{itemize}
    \item \textbf{Step I. local calculation.} For the $k$-th source, we update the parameter based on gradient descent:
    \begin{equation}
        \overline{\boldsymbol{\beta}}^{(k)\{t+1\}}_{b}=\widehat{\boldsymbol{\beta}}^{(k)\{t\}}_{b}-\omega \mathbf{g}^{(k)\{t+1\}}_{b},
    \end{equation}
    where $\omega$ is the learning rate, and $\mathbf{g}^{(k)\{t+1\}}_{b}$ denotes the gradient of the loss function, calculated by 
    \begin{equation}
    \label{eq:gradient}  
    \mathbf{g}^{(k)\{t+1\}}_{b}=-\frac{1}{N_{b}}\frac{\partial \widehat{\mathcal{L}}^{(k)}_{b}\big(\overline{\boldsymbol{\beta}}^{(k)\{t\}}_{b}\big)}{\partial {\overline{\boldsymbol{\beta}}}^{(k)\{t\}}_{b}}= -\frac{1}{{N}_{b}}\left[\mathbf{U}\big(\overline{\boldsymbol{\beta}}^{(k)\{t+1\}}_{b}; \mathcal{D}_b^{(k)}\big)+ \widehat{\mathbf{J}}^{(k)}_{b-1}\big(\overline{\boldsymbol{\beta}}^{(k)\{t+1\}}_{b}-\widehat{\boldsymbol{\beta}}^{(k)}_{b-1}\big) \right].
\end{equation}

    After the calculation, we transmit the updated model parameter $\overline{\boldsymbol{\beta}}^{(k)\{t+1\}}_{b}$ to the master server.

    \item \textbf{Step II. master aggregation.} The master server aggregates all the updated model parameters $\overline{\mathbf{B}}_{b}^{\{t+1\}}=\big(\overline{\boldsymbol{\beta}}^{(1)\{t+1\}}_{b},\dots,\overline{\boldsymbol{\beta}}^{(K)\{t+1\}}_{b}\big)$, and use the proximal operator to get the output for the $(t+1)$-th iteration:
    \begin{equation}
        \label{eq:proximal_update}  
        \widehat{\mathbf{B}}^{\{t+1\}}_{b} = \mathcal{P}_{a}\left(\overline{\mathbf{B}}_{b}^{\{t+1\}}; \lambda_{1,b},\lambda_{2,b} \right),
    \end{equation}
    where the function $\mathcal{P}_{a}(\overline{\mathbf{B}}_{b}^{\{t+1\}}; \lambda_{1,b},\lambda_{2,b})$ denotes the proximal operator associated with the penalty terms in (\ref{eq:obj}), which is defined as:
    {\small
    \begin{equation}
        \label{eq:prox}
        \mathcal{P}_{a}(\overline{\mathbf{B}}_{b}^{\{t+1\}}; \lambda_{1,b},\lambda_{2,b})=\arg\min_{\mathbf{B}} \frac{1}{2}\left\|\mathbf{B}-\overline{\mathbf{B}}_{b}^{\{t+1\}}\right\|_{F}^{2} + \sum_{k=1}^{K} \sum_{j=1}^{p} \rho_{a}\left(\left|\beta^{(k)}_{j}\right|, \lambda_{1,b}\right) + \sum_{k_1< k_2} \rho_{a}\left(\left\|\boldsymbol{\beta}^{(k_1)}-\boldsymbol{\beta}^{(k_2)}\right\|_2, \lambda_{2,b}\right),
    \end{equation}
    }
    where $\|\cdot\|_{F}$ denotes the Frobenius norm. We use the alternating direction method of multipliers (ADMM) \citep{boyd_distributed_2011} to solve the optimization problem \eqref{eq:prox}. The detailed calculation procedures are presented in Section S1.1 of the supplementary material.
\end{itemize}

Overall, we optimize \eqref{eq:obj} by repeating Steps I--II until convergence. The convergence properties for proximal gradient descent have been widely investigated in the literature \citep{li_convergence_2017}, and in most of our numerical studies, the convergence can be achieved within 50 iterations. The detailed procedures are outlined in Algorithm S1 of the supplementary material. The objective function includes three tuning parameters: $\lambda_{1,b}$, $\lambda_{2,b}$, and $a$. In our numerical study, we set $a=3$ \citep{zhang_nearly_2010} and perform a grid search with modified Bayesian information criteria (mBIC; \citealp{ma_concave_2017}) to select $\lambda_{1,b}$ and $\lambda_{2,b}$. Further details on mBIC calculations can be found in Section S1.2 of the supplementary material.  

\section{Theoretical properties}\label{sec:Theory}
\subsection{Notations and Definitions}
We introduce the following essential definitions.
For the $u$-th batch from source $k$, let $\mathbf{X}^{(k)}_{u}=(\mathbf{x}_{u,1}^{(k)},\dots,\mathbf{x}_{u,n_{u,k}}^{(k)})^\top$ be the design matrix and $\mathbf{Y}^{(k)}_{u}=(y_{u,1}^{(k)},\dots,y_{u,n_{u,k}}^{(k)})^\top$ be the response vector.
We define $\bm\varphi_u^{(k)}=\mathbf{X}^{(k)}_{u}\bm\beta^{(k)}$ and let $\mathbf{d}(\bm\varphi_u^{(k)})=\big[d(\varphi_{u,1}^{(k)}),\dots,d(\varphi_{u,n_{u,k}}^{(k)})\big]^{\top}$ be a vector function of $\bm\varphi_u^{(k)}$.
For simplicity, let $\boldsymbol{\mu}\big(\bm\varphi_u^{(k)}\big)=\mathbf{d}'(\bm\varphi_u^{(k)})$ and $\boldsymbol{\Sigma}\big(\bm\varphi_u^{(k)}\big)=\text{diag}\{\mathbf{d}''(\bm\varphi_u^{(k)})\}$. 
Let $|\mathcal{G}_{\min}|=\min_{g\in [G]}|\mathcal{G}^{(g)}|$ denote the minimal subgroup size. Define ${n}^{(g)}_{u}=\sum_{k\in \mathcal{G}^{(g)}}n_{u,k}, u\in[b]$ as the sample size of aggregated sources in subgroup $g$ for the $u$-th batch, and ${N}^{(g)}_{b}=\sum_{k\in \mathcal{G}^{(g)}}N_{b,k}=\sum_{u=1}^{b}n_{u}^{(g)}$ as the cumulative sample size up to batch $b$. We denote $\bar{N}_b=\min_{g\in[G]} {N}^{(g)}_{b}$.  
Let $\mathbf{B}^{*}=(\boldsymbol{\beta}^{*(1)},\dots,\boldsymbol{\beta}^{*(K)})$ denote the true coefficient matrix corresponding to $\mathbf B$. Define $\mathcal B_k=\{j\in[p]:\bm\beta_j^{*(k)}\ne 0\}$ as the support of $\bm\beta^{*(k)}$, with $\mathcal B_k^c$ denoting its complement.
Accordingly, let $\boldsymbol{\Xi}=(\boldsymbol{\xi}^{(1)},\dots, \boldsymbol{\xi}^{(G)})$ be the distinct coefficient matrix, where $\bm\xi^{(g)}=\boldsymbol{\beta}^{(k)}$ for any $k\in\mathcal G^{(g)}$, and $\boldsymbol{\Xi}^{*}=(\boldsymbol{\xi}^{*(1)},\dots, \boldsymbol{\xi}^{*(G)})$ be the corresponding true coefficient matrix. We define $\mathcal A_g=\mathcal B_{k}$ for any $k\in\mathcal G^{(g)}$ as the index set for significant variables in $g$-th subgroup, $s_g=|\mathcal A_g|$ as the corresponding cardinality. Denote $s=\max_{g} s_{g}$. 

Consider the following two subspaces of $\mathbb R^{p\times K}$:
\begin{equation}
\label{eq:space}
\begin{aligned}
 \mathcal{M}_{\mathcal{G}}&=\left\{\mathbf{B}\in \mathbb{R}^{p\times K}:\boldsymbol{\beta}^{(k_1)}=\bm\beta^{(k_2)},\ \text{for any}\ k_1,k_2\in \mathcal{G}^{(g)}, g\in[G] \right\},\\
 \mathcal{M}_{\mathcal{B}}&=\left\{\mathbf{B}\in \mathbb{R}^{p\times K}:\boldsymbol{\beta}_{\mathcal B_k^c}^{(k)}=\bm 0_{p-s_g},\ \text{for any}\ k\in\mathcal{G}^{(g)},g\in[G] \right\}. 
 \end{aligned}
 \end{equation}
When the true subgroup structure $\mathcal G=\{\mathcal G^{(1)},\dots,\mathcal G^{(G)}\}$ and the sparsity structure $\{\mathcal A_g:g\in[G]\}$ are known, the ``subgroup-sparsity-oracle" estimator for $\mathbf B$ (up to batch $b$) is defined as
\begin{equation}
    \label{eq:obj_or_1}
    \begin{aligned}
        \widehat{\mathbf{B}}^{or}_{b}=\arg\min_{\mathbf{B}\in \mathcal M_{\mathcal G}\cap\mathcal M_{\mathcal B}} \left\{\widehat{\mathcal{L}}_{b}(\mathbf{B})=-\frac{1}{{N}_{b}}\sum_{k=1}^{K} \widehat{\mathcal{L}}^{(k)}_{b}\big(\boldsymbol{\beta}^{(k)}\big)\right\} .
    \end{aligned}
\end{equation}
It can be shown that, minimizing $\widehat{\mathcal{L}}_{b}(\mathbf{B})$
under the feasible set $\mathcal M_{\mathcal G}\cap\mathcal M_{\mathcal B}$ is equivalent to the constrained minimization of $\boldsymbol{\Xi}$ in that
\begin{equation}\label{eq:obj_or_online_sep}  
\widehat{\boldsymbol{\Xi}}_{b}^{or}=\arg\min_{\bm\Xi\in\mathcal M_{\mathcal A}^{or}}\left\{\widehat{\mathcal{L}}_{b}^{or}(\boldsymbol{\Xi})=-{N}_{b}^{-1}\sum_{g\in[G]}\sum_{k\in \mathcal{G}^{(g)}}\widehat{\mathcal{L}}^{(k)}_{b}\big(\boldsymbol{\xi}^{(g)}\big)\right\},
\end{equation}
where $\mathcal{M}_{\mathcal{A}}^{or}=\left\{\bm{\Xi}\in \mathbb{R}^{p\times G}:\bm\xi_{\mathcal A_g^c}^{(g)}=\bm 0_{p-s_g},g\in[G] \right\}$.

\subsection{Asymptotic properties}

In our theory, we consider a scenario where the sample size for each batch, denoted as $n_{u,k}$, is divergent. For simplicity, we assume that $n_{u,k}\asymp n_{u,*}$ for $u\in [b]$ and $k\in [K]$, implying $N_{u,k}\asymp N_{u,*}$. Furthermore, the number of variables ($p$), important variables ($s$), and sources ($K$) can all go to infinity. To establish the asymptotic properties, we outline some regularity conditions in the following. 

\begin{Condition}\label{cond:bound}
    Data $\big\{(\mathbf{x}_{u,i}^{(k)},y_{u,i}^{(k)}):k\in[K],u\in[b],i\in [n_{u,k}]\big\}$ are independent and identically distributed. 
    Moreover, there exists a constant $C_x>0$ such that
    \begin{equation}
        \begin{array}{c}
             \max_{k\in[K],u\in[b]}\big\|{\mathbf X}_u^{(k)}\big\|_{\max}\le C_{x}, \\   \sigma_{\max}\left[\mathbf{X}^{(k)\top}_{u,\mathcal{A}_{g}}\mathbf{X}^{(k)}_{u,\mathcal{A}_{g}} \right]\leq C_{x} n_{u,k}, \quad k\in \mathcal{G}^{(g)}, g\in [G],
        \end{array}
    \end{equation}
    where $\|\cdot\|_{\max}$ and $\sigma_{\max}(\cdot)$ denote the maximal absolute value and the maximum eigenvalue of the matrix.
\end{Condition}

\begin{Condition} \label{cond:minimal_signal}
    Define the minimal signal level ${d}_{1}=\min_{g\in[G],j\in\mathcal A_g,k\in \mathcal{G}^{(g)}}|\beta_{j}^{*(k)}|$ and the minimal distance between different subgroups ${d}_{2}=\min_{g_1\neq g_2\in[G],k_1\in \mathcal{G}^{(g_1)}, k_2\in \mathcal{G}^{(g_2)}}\|\boldsymbol{\beta}^{*(k_1)}-\boldsymbol{\beta}^{*(k_2)}\|_2$. For $u\in [b]$, assume ${d}_{1}>2a\lambda_{1,u}$, $d_2> 2a\lambda_{2,u}$, and $\lambda_{1,u}\gg \sqrt{[\log (pK)+s]/{\bar{N}_{u}}}$, $\lambda_{2,u}\gg \sqrt{{s}/{\bar{N}_{u}}}$.
\end{Condition}

\begin{Condition} \label{cond:design_matrix1}
    For the $g$-th ($g\in [G]$) subgroup, define the neighborhood around the vector of true coefficients $\boldsymbol{\xi}^{*(g)}_{\mathcal{A}_{g}}$ as $\mathcal{N}_{0}^{(g)}=\big\{\boldsymbol{\xi}_{\mathcal{A}_{g}}\in \mathbb{R}^{s_{g}}: \|\boldsymbol{\xi}_{\mathcal{A}_{g}}-\boldsymbol{\xi}^{*(g)}_{\mathcal{A}_{g}}\|_{\infty}\leq d_1/2\big\}$. Then for each $k\in \mathcal{G}^{(g)}$ and batch $u=1,\dots,b$, we assume that
    \begin{gather}
        \min_{\boldsymbol{\alpha}\in \mathcal{N}_{0}^{(g)}} \sigma_{\min}\left[\mathbf{X}^{(k)\top}_{u,\mathcal{A}_{g}}\boldsymbol{\Sigma}\big(\mathbf{X}^{(k)}_{u,\mathcal{A}_{g}}\boldsymbol{\alpha} \big)\mathbf{X}^{(k)}_{u,\mathcal{A}_{g}} \right]\geq c_1 n_{u,k},
        \label{eq:minimal_eigenvalue}   \\
        \max_{\boldsymbol{\alpha}\in \mathcal{N}_{0}^{(g)}} \sigma_{\max}\left[\mathbf{X}^{(k)\top}_{u,\mathcal{A}_{g}}\boldsymbol{\Sigma}\big(\mathbf{X}^{(k)}_{u,\mathcal{A}_{g}}\boldsymbol{\alpha} \big)\mathbf{X}^{(k)}_{u,\mathcal{A}_{g}} \right]\leq C_1 n_{u,k},  
        \label{eq:maximal_eigenvalue} \\
        \max_{\boldsymbol{\alpha}\in \mathcal{N}_{0}^{(g)}} \max_{j=1}^{p} \sigma_{\max}\left[ \mathbf{X}^{(k)\top}_{u,\mathcal{A}_{g}}\text{diag}\left\{|\mathbf{X}^{(k)\top}_{u, \cdot j}|\circ \mathbf{d}'''\big(\mathbf{X}^{(k)}_{u,\mathcal{A}_{g}}\boldsymbol{\alpha} \big) \right\} \mathbf{X}^{(k)}_{u,\mathcal{A}_{g}}\right] \leq C_2 n_{u,k},
        \label{eq:large_deviance}
    \end{gather}
    where $c_1$, $C_1$ and $C_2$ are three positive constants, $\sigma_{\min}(\cdot)$ denotes the minimal eigenvalue of the matrix.
\end{Condition}  

\begin{Condition}
    \label{cond:tr}  
    For each subgroup $g\in [G]$ and batch $u\in [b]$, it satisfies that for any $n_{u}^{(g)}$-dimensional vector $\boldsymbol{\eta}$ and $\epsilon>0$,
    \begin{equation}\label{prob_bound}
        \Pr\left[\left|\boldsymbol{\eta}^{\top} \left\{\widetilde{\mathbf{Y}}^{(g)}_{u}-\boldsymbol{\mu}\big(\widetilde{\mathbf{X}}^{(g)}_{u} \boldsymbol{\xi}^{*(g)}\big) \right\} \right|<\|\boldsymbol{\eta}\|_2\epsilon \right]>1-2\exp\left[-\kappa \epsilon^2 \right],
    \end{equation}
    where $\widetilde{\mathbf{X}}_{u}^{(g)}=\big({\mathbf{X}}_{u}^{(k)\top},k\in\mathcal G^{(g)}\big)^{\top}$, $\widetilde{\mathbf{Y}}_{u}^{(g)}=\big({\mathbf{Y}}_{u}^{(k)\top},k\in\mathcal G^{(g)}\big)^{\top}$, and $\kappa>0$ is a constant.
\end{Condition}

\begin{Condition}\label{cond:design_matrix2}
    For any $g$-th ($g\in [G]$) subgroup, we assume that  
    \begin{gather}
        \left\|\mathbf{X}^{(k)\top}_{u,\mathcal{A}_{g}^{c}} \boldsymbol{\Sigma}\big(\mathbf{X}^{(k)}_{u}\boldsymbol{\beta}^{*(k)} \big) \mathbf{X}^{(k)}_{u,\mathcal{A}_{g}}\right\|_{2,\infty}\leq C_3 n_{u,k},\quad k\in \mathcal{G}^{(g)}, u\in [b],
        \label{eq:re_21} 
    \end{gather}
    where $C_3$ is a positive constant. For matrix $\mathbf{H}$, $\|\mathbf{H}\|_{2,\infty}=\max_{\mathbf x\in\mathbb R^p,\|\mathbf{x}\|_{2}=1}\|\mathbf{H}\mathbf{x}\|_{\infty}$.
\end{Condition}

\begin{Condition}\label{cond:penalty} 
The penalty function $\rho_{a}(x, \lambda)$ is a symmetric function of $x$ with $\rho_{a}(0, \lambda)=0$. It is non-decreasing and concave in $[0,+\infty)$. Its first-order derivative $\rho'_{a}(x, \lambda)$ exists and is continuous for $x\in [0,\infty)$, $\rho_{a}'(0+, \lambda)=\lambda$.  $\rho_{a}(x, \lambda)$ equals to a constant $C_{\lambda,a}$ for $|x| \geq a\lambda$, i.e., $\rho'_{a}(x, \lambda)=0$ for $|x| \geq a\lambda$. 
\end{Condition}

The detailed explanation for these conditions is summarized in the Supplementary Material Section S2. Based on the conditions, we can develop the following theoretical properties.

\begin{theorem}\label{lemma1}
    Consider the online estimate in the first batch ($b=1$). Under Conditions \ref{cond:bound} -- \ref{cond:penalty}, we have that 
    \begin{enumerate}[(1)]
    \item The ``subgroup-sparsity-oracle" common coefficient estimator defined in (\ref{eq:obj_or_online_sep}) satisfies: 
    $$
    \max_{g\in[G]}\Big\|\widehat{\boldsymbol{\xi}}^{or(g)}_{1, \mathcal{A}_{g}}-\boldsymbol{\xi}^{*(g)}_{\mathcal{A}_{g}}\Big\|_2=O_p\Big(\sqrt{{s}/{\bar{N}_{1}}}\Big),
    $$
    which implies that the ``subgroup-sparsity-oracle" estimator defined in (\ref{eq:obj_or_1}) also satisfies that  
    $
    \max_{k\in[K]}\left\|\widehat{\boldsymbol{\beta}}^{or(k)}_{1, \mathcal{B}_{k}}-\boldsymbol{\beta}^{*(k)}_{\mathcal{B}_{k}}\right\|_2=O_p\Big(\sqrt{{s}/{\bar{N}_{1}}}\Big)
    $.
    \item There exists a strictly local minimizer $\widehat{\mathbf{B}}_{1}$ of $\mathcal{Q}_1^{\text{on}}(\mathbf{B})$ that
    $
\mathrm{Pr}\Big[\widehat{\mathbf{B}}_{1}=\widehat{\mathbf{B}}^{or}_{1}\Big]\to 1.
    $
    \end{enumerate}
\end{theorem}
Theorem \ref{lemma1} establishes the consistency of the estimated coefficients for the first batch ($b=1$) based on the offline estimate. Result (1) shows the consistency of the ``subgroup-sparsity-oracle" estimator for common subgroup coefficients $\widehat{\mathbf B}_1^{or}$, that is, $\max_{k\in[K]}\big\|\widehat{\boldsymbol{\beta}}^{or(k)}_{1, \mathcal{B}_{k}}-\boldsymbol{\beta}^{*(k)}_{\mathcal{B}_{k}}\big\|_2$ converges at a rate of $\sqrt{{s}/{\bar{N}_{1}}}$. Result (2) further indicates that $\widehat{\mathbf{B}}^{or}_{1}$ is a strictly local minimizer of $\mathcal Q_1^{\mathrm{on}}(\mathbf B)$ in (\ref{eq:obj}) with probability approaching 1. Together, these results imply that $\widehat{\mathbf{B}}_{1}$ is asymptotically equivalent to $\widehat{\mathbf{B}}^{or}_{1}$ and then inherits the same convergence rate. Consequently, since both variable selection and clustering rely on this structure, we conclude that $\mathrm{Pr}[\cap_{k=1}^K\{\widehat{\mathcal B}_k=\mathcal B_k\}]\to 1$ and $\mathrm{Pr}[\widehat{\mathcal G}=\mathcal G]\to 1$, indicating consistency in variable selection and clustering.

\begin{theorem}\label{theorem2}
Consider the online estimate in the $b$-th batch ($b\ge 2$). Under Conditions \ref{cond:bound} -- \ref{cond:penalty}, we have that 
\begin{enumerate}[(1)]
    \item The ``subgroup-sparsity-oracle" common coefficient estimator defined in (\ref{eq:obj_or_online_sep}) satisfies:
\[
\max_{g\in[G]}\Big\|\widehat{\boldsymbol{\xi}}^{or(g)}_{b,\mathcal{A}_{g}}-\boldsymbol{\xi}^{*(g)}_{\mathcal{A}_{g}}\Big\|_2=O_p\Big(\sqrt{{s}/{\bar{N}_{b}}}\Big),
\]
which directly implies the ``subgroup-sparsity-oracle" estimator  defined in (\ref{eq:obj_or_1}) also satisfies that  
$
\max_{k\in[K]}\left\|\widehat{\boldsymbol{\beta}}^{or(k)}_{b, \mathcal{B}_{k}}-\boldsymbol{\beta}^{*(k)}_{\mathcal{B}_{k}}\right\|_2=O_p\Big(\sqrt{{s}/{\bar{N}_{b}}}\Big)
$.
    \item There exists a strictly local minimizer $\widehat{\mathbf{B}}_{b}$ of $\mathcal{Q}_{b}^{\text{on}}(\mathbf{B})$ that
    $
    \mathrm{Pr}\Big[\widehat{\mathbf{B}}_{b}=\widehat{\mathbf{B}}^{or}_{b}\Big]\to 1.
    $
\end{enumerate}
\end{theorem}
Theorem \ref{theorem2} establishes the consistency of online estimates using results from Theorem \ref{lemma1} and an inductive approach. 
Similar to Theorem \ref{lemma1}, the estimator $\widehat{\mathbf{B}}_{b}$ from the $b$-th batch converges at the rate: $\max_{k\in[K]}\big\|\widehat{\boldsymbol{\beta}}^{(k)}_{b, \mathcal{B}_{k}}-\boldsymbol{\beta}^{*(k)}_{\mathcal{B}_{k}}\big\|_2=O_p\big(\sqrt{{s}/{\bar{N}_{b}}}\big)$, ensuring consistency in variable selection and clustering. Compared to $\widehat{\mathbf{B}}_{1}$ from the first batch, $\widehat{\mathbf{B}}_{b}$ converges faster due to the larger sample size $\bar N_b$, matching the convergence rate of the offline estimator. 
Therefore, Theorem \ref{theorem2} indicates that $\widehat{\mathbf{B}}_{b}$ is asymptotically equivalent to $\widetilde{\mathbf B}_b$, consistent with the online learning literature \citep{luo_renewable_2020}. However, the high-dimensional, multi-source data settings introduce unique theoretical challenges. Additionally, the consistency in clustering shares similarities with integrative analysis and federated clustering literature. To the best of our knowledge, this property has not been explored in online learning. Thus, our investigation provides unique contributions to this field. 

\section{Simulation Studies}\label{sec:simulation}
In this section, we conduct simulations to evaluate the finite sample performance of the proposed method. The data generation process is as follows: (a) We consider scenarios with $K \in \{8, 16, 32\}$ sources, with streaming data arriving in 10 batches. Each source starts with 100 subjects ($n_{1,k}=100$ for $k\in [K]$), followed by $n_{u,k} \in \{40, 80\}$ subjects in subsequent batches ($u=2,\dots,10$). (b) The number of covariates is set at $p \in \{50, 100, 200\}$. The covariates are generated from a multivariate normal distribution with a mean of zero and an autoregressive covariance structure: $\mathbf{x}_{u,i}^{(k)} \sim N(\mathbf{0}_p, \boldsymbol{\Sigma}_p)$, where $[\boldsymbol{\Sigma}_p]_{i,j}=0.5^{|i-j|}$. (c) The response variable $y_{u,i}^{(k)}$ is randomly drawn from the distribution $f(y|\mathbf{x}_{u,i}^{(k)},\boldsymbol{\beta}^{(k)})$. 
We use logistic and linear regression analysis in our simulations.

We explore three examples to illustrate various subgroup structures among data sources. Specifically, Example 1 depicts a homogeneous scenario in which all sources share the same model. In contrast, Examples 2 and 3 present heterogeneous scenarios, with models for different sources grouped into 2 and 4 subgroups, respectively.

\noindent\textbf{Example 1.} In this homogeneous case, we have $\boldsymbol{\beta}^{*(k)}=\boldsymbol{\xi}^{*}$ for $k=1,\dots, 8$. We randomly select two disjoint subsets of variables, $\mathcal{C}_{1}$ and $\mathcal{C}_2$, each containing 4 variables. The coefficient vector $\boldsymbol{\xi}^{*}$ is defined as: $\boldsymbol{\xi}_{\mathcal{C}_1}^{*}=0.6\mathbf{1}_4$, $\boldsymbol{\xi}_{\mathcal{C}_2}^{*}=-0.6\mathbf{1}_4$ and $\boldsymbol{\xi}^{*}_{\mathcal{C}_1^c\cap \mathcal{C}_2^c}=\mathbf{0}_{p-8}$.

\noindent\textbf{Example 2.} In this case, coefficients from different sources are divided into two subgroups: $\boldsymbol{\beta}^{*(k)}=\boldsymbol{\xi}^{*(1)}$ for $k=1,\dots, 4$ and $\boldsymbol{\beta}^{*(k)}=\boldsymbol{\xi}^{*(2)}$ for $k=5,\dots, 8$, with $\boldsymbol{\xi}^{*(1)}\neq \boldsymbol{\xi}^{*(2)}$. The index sets $\mathcal{C}_1$ and $\mathcal{C}_2$ are defined similarly to Example 1, with coefficient vectors given by: (i) $\boldsymbol{\xi}^{(1)*}_{\mathcal{C}_1}=0.6\mathbf{1}_4$, $\boldsymbol{\xi}^{(1)*}_{\mathcal{C}_2}=-0.6\mathbf{1}_4$, $\boldsymbol{\xi}^{(1)*}_{\mathcal{C}_1^c\cap \mathcal{C}_2^c}=\mathbf{0}_{p-8}$; and (ii) $\boldsymbol{\xi}^{(2)*}_{\mathcal{C}_1}=-0.6\mathbf{1}_4$, $\boldsymbol{\xi}^{(2)*}_{\mathcal{C}_2}=0.6\mathbf{1}_4$, $\boldsymbol{\xi}^{(2)*}_{\mathcal{C}_1^c\cap \mathcal{C}_2^c}=\mathbf{0}_{p-8}$.

\noindent\textbf{Example 3.} Here, coefficients from different sources are divided into four subgroups: $\boldsymbol{\beta}^{*(k)}=\boldsymbol{\xi}^{*(1)}$ for $k=1,2$, $\boldsymbol{\beta}^{*(k)}=\boldsymbol{\xi}^{*(2)}$ for $k=3,4$, $\boldsymbol{\beta}^{*(k)}=\boldsymbol{\xi}^{*(3)}$ for $k=5,6$, and $\boldsymbol{\beta}^{*(k)}=\boldsymbol{\xi}^{*(4)}$ for $k=7,8$, with $\boldsymbol{\xi}^{(1)*}\neq \boldsymbol{\xi}^{*(2)}\neq \boldsymbol{\xi}^{*(3)}\neq \boldsymbol{\xi}^{*(4)}$. The index sets $\mathcal{C}_1$ and $\mathcal{C}_2$ are defined as in Example 1, with the following coefficient vectors: (i) $\boldsymbol{\xi}^{(1)*}_{\mathcal{C}_1}=0.6\mathbf{1}_4$, $\boldsymbol{\xi}^{(1)*}_{\mathcal{C}_2}=0.6\mathbf{1}_4$, $\boldsymbol{\xi}^{(1)*}_{\mathcal{C}_1^c\cap \mathcal{C}_2^c}=\mathbf{0}_{p-8}$; (ii) $\boldsymbol{\xi}^{(2)*}_{\mathcal{C}_1}=-0.6\mathbf{1}_4$, $\boldsymbol{\xi}^{(2)*}_{\mathcal{C}_2}=0.6\mathbf{1}_4$, $\boldsymbol{\xi}^{(2)*}_{\mathcal{C}_1^c\cap \mathcal{C}_2^c}=\mathbf{0}_{p-8}$; (iii) $\boldsymbol{\xi}^{(1)*}_{\mathcal{C}_1}=0.6\mathbf{1}_4$, $\boldsymbol{\xi}^{(1)*}_{\mathcal{C}_2}=-0.6\mathbf{1}_4$, $\boldsymbol{\xi}^{(1)*}_{\mathcal{C}_1^c\cap \mathcal{C}_2^c}=\mathbf{0}_{p-8}$; and (iv) $\boldsymbol{\xi}^{(2)*}_{\mathcal{C}_1}=-0.6\mathbf{1}_4$, $\boldsymbol{\xi}^{(2)*}_{\mathcal{C}_2}=-0.6\mathbf{1}_4$, $\boldsymbol{\xi}^{(2)*}_{\mathcal{C}_1^c\cap \mathcal{C}_2^c}=\mathbf{0}_{p-8}$.

In addition to the proposed approach, we examine several relevant alternatives: (a)  {Oracle}: This is an offline federated learning method, which assumes that we have simultaneous access to all data batches. The objective function for Oracle is similar to the proposed method, except that no approximation is adopted for the log-likelihood function. Although impractical for data streaming scenarios, Oracle serves as a benchmark for comparison.
(b)  {Ind}: This is the individualized online learning method. It updates the model in each source independently while 
ignoring potential connections between them. (c)  {Homo}: This is the online divide-and-conquer method, which assumes homogeneity among sources. It first updates each data source, similar to Ind, and then aggregates the results to get a global estimation.

We evaluate these methods based on variable selection, estimation, prediction, and clustering. For variable selection, we use the average true positive rate (TPR) and false positive rate (FPR) across all sources. Estimation performance is assessed using the sum of squared errors $\text{SSE}=(1/K)\sum_{k=1}^{K}\|\widehat{\boldsymbol{\beta}}^{(k)}_{u}-\boldsymbol{\beta}^{*(k)}\|_2^2$. For prediction, we generate 2000 independent test samples for each source and evaluate performance with mean squared error (MSE) for linear regression and area under curve (AUC) for logistic regression. 
For clustering, we analyze the number of identified subgroups ($\widehat{G}$) and the adjusted rand index (ARI), which measures the agreement between the identified and true cluster structures, ranging from -1 to 1 (higher values indicate better agreement). Note that Ind and Homo cannot cluster data sources, so their clustering performance is not evaluated. 

The simulation results from 100 replicates are reported in Table \ref{tab:logistic_exp2}, together with additional tables and figures in Section S3 of the supplementary material. Observations are similar across settings. In the homogeneous example (Example 1), the proposed method effectively identifies the underlying subgroup structure for data sources and improves performance in variable selection, estimation, and prediction. It outperforms the Ind method and demonstrates the advantages of our joint learning strategy. 
The proposed method slightly outperforms the Homo method and is comparable to the infeasible Oracle method, also indicating its effectiveness. Similar patterns can be observed in heterogeneous scenarios (Examples 2 and 3).
Figures S1--S3 in the supplementary material show the trace plots of AUC (logistic regression) and MSE (linear regression), as well as the SSE for coefficient estimation across batches. 
Overall, the performance of Ind and the proposed method improves with more batches, and the proposed method generally outperforms Ind, underscoring the competitive performance of the proposed estimator.  

\begin{table}
\TABLE
{Simulation results for Example 2 with the logistic regression model, using $K=8$ data sources. Each cell displays the mean (sd) based on 100 replicates. \label{tab:logistic_exp2}}
{
\begin{tabular}{ccccccccc}
\toprule
$p$ & $n_{u,k}$ & Method &TPR & FPR & SSE & AUC & $\widehat{G}$ & ARI \\
\midrule 
\multirow{8}{*}{50}  & \multirow{4}{*}{40} & Oracle & 1.00(0.00) & 0.00(0.01) & 0.036(0.012) & 0.826(0.020) & 2.0(0.0) & 1.00(0.00) \\
                     &                     & Ind    & 0.61(0.09) & 0.01(0.01) & 1.287(0.232) & 0.761(0.035) & -        & -          \\
                     &                     & Homo   & 0.99(0.03) & 0.10(0.05) & 2.976(0.041) & 0.501(0.003) & -        & -          \\
                     &                     & Proposed   & 0.95(0.06) & 0.00(0.01) & 0.227(0.189) & 0.818(0.023) & 2.0(0.2) & 0.99(0.05) \\
                     \cmidrule{2-9}
                     & \multirow{4}{*}{80} & Oracle & 1.00(0.00) & 0.00(0.00) & 0.024(0.013) & 0.828(0.012) & 2.0(0.0) & 1.00(0.00) \\
                     &                     & Ind    & 0.71(0.07) & 0.01(0.01) & 0.968(0.203) & 0.783(0.022) & -        & -          \\
                     &                     & Homo   & 1.00(0.00) & 0.06(0.04) & 2.947(0.042) & 0.499(0.004) & -        & -          \\
                     &                     & Proposed   & 0.99(0.02) & 0.00(0.00) & 0.069(0.093) & 0.826(0.014) & 2.0(0.0) & 1.00(0.00) \\
                     \midrule  
\multirow{8}{*}{100} & \multirow{4}{*}{40} & Oracle & 1.00(0.00) & 0.00(0.00) & 0.036(0.011) & 0.825(0.014) & 2.0(0.0) & 1.00(0.00) \\
                     &                     & Ind    & 0.51(0.09) & 0.01(0.00) & 1.574(0.258) & 0.742(0.031) & -        & -          \\
                     &                     & Homo   & 0.94(0.10) & 0.05(0.02) & 2.972(0.046) & 0.503(0.005) & -        & -          \\
                     &                     & Proposed   & 0.90(0.06) & 0.00(0.00) & 0.437(0.198) & 0.808(0.019) & 2.0(0.0) & 1.00(0.00) \\
                     \cmidrule{2-9}
                     & \multirow{4}{*}{80} & Oracle & 1.00(0.00) & 0.00(0.00) & 0.021(0.008) & 0.827(0.014) & 2.0(0.0) & 1.00(0.00) \\
                     &                     & Ind    & 0.65(0.07) & 0.00(0.00) & 1.150(0.207) & 0.771(0.024) & -        & -          \\
                     &                     & Homo   & 1.00(0.00) & 0.02(0.02) & 2.942(0.025) & 0.499(0.005) & -        & -          \\
                     &                     & Proposed   & 0.97(0.05) & 0.00(0.00) & 0.148(0.146) & 0.822(0.016) & 2.0(0.0) & 1.00(0.00) \\
                     \midrule  
\multirow{8}{*}{200} & \multirow{4}{*}{40} & Oracle & 1.00(0.00) & 0.00(0.00) & 0.035(0.018) & 0.829(0.014) & 2.0(0.0) & 1.00(0.00) \\
                     &                     & Ind    & 0.48(0.10) & 0.00(0.00) & 1.658(0.279) & 0.737(0.029) & -        & -          \\
                     &                     & Homo   & 0.97(0.08) & 0.03(0.01) & 2.951(0.033) & 0.499(0.002) & -        & -          \\
                     &                     & Proposed   & 0.87(0.07) & 0.00(0.00) & 0.554(0.187) & 0.806(0.017) & 2.0(0.0) & 1.00(0.00) \\
                     \cmidrule{2-9}
                     & \multirow{4}{*}{80} & Oracle & 1.00(0.00) & 0.00(0.00) & 0.025(0.009) & 0.822(0.019) & 2.0(0.0) & 1.00(0.00) \\
                     &                     & Ind    & 0.52(0.10) & 0.00(0.00) & 1.521(0.277) & 0.742(0.043) & -        & -          \\
                     &                     & Homo   & 0.94(0.11) & 0.01(0.01) & 2.917(0.019) & 0.497(0.004) & -        & -          \\
                     &                     & Proposed   & 0.86(0.10) & 0.00(0.00) & 0.482(0.304) & 0.803(0.031) & 2.0(0.0) & 1.00(0.00)
\\
\bottomrule
\end{tabular}
} {}
\end{table}

\section{Real Data Application}\label{sec:realdata}
\subsection{Analysis of financial platform data}


We apply the proposed method to financial data from a peer-to-peer (P2P) lending platform to predict the success of credit-accredited bids.
On this platform, users seeking funding post bids containing personal details (e.g., age, gender, marital status), bid specifics (e.g., total funds sought, borrowing period, interest rate), reasons for borrowing, and other relevant information. 
Investors use this data to determine whether and how much to invest. 
A bid is labeled as ``successful" if it secures sufficient funding; otherwise, it is marked as ``lost". 
Predicting the success or failure of new bids based on user and bid information is crucial for the P2P platform, as it can highlight key factors influencing successful fundraising and offer valuable insights into the platform's operations and growth. 

The dataset contains credit-accredited bid information from $K=13$ provinces (data sources) in China, collected over 13 days from September 7 to September 16, 2020, totaling 8,631 observations.
Table S9 presents the sample sizes from various sources and batches.
Sample sizes vary by source: Guangdong has the most observations, with 1,441 observations (48 to 79 per batch), while Jiangxi has the least at 489 (31 to 46 per batch).
In contrast, the sample sizes are relatively balanced between different batches for each source.  
The response variable denotes bid status, with $y_i=1$ for successful bids and $y_i=0$ otherwise, resulting in an overall success rate of 4.83\%. 
After applying feature engineering techniques such as one-hot encoding for categorical variables and log transformation or standardization for continuous variables, as well as feature interaction generation, we finally generate $p=57$ predictors. 
These predictors provide valuable information on both borrowers and bids, significantly influencing investors' decisions and bid outcomes.

After applying the proposed method to this dataset via a logistic model, we identified three subgroups of provinces: the first subgroup includes Guangdong, Fujian, Hunan, Hebei, and Jiangxi (5 provinces); the second subgroup comprises Zhejiang, Shandong, Hubei, Sichuan, Henan, Anhui, and Guangxi (7 provinces); and the third consists solely of Beijing (1 province).
Table \ref{tab:real_data_estimation} presents the results of variable selection and coefficient estimates for these subgroups, revealing 13, 10, and 10 significant variables, respectively, with substantial overlap and consistent coefficient signs across all models, though their magnitudes vary. 
Key factors influencing successful fundraising include smaller bid amounts, which align with expectations, while higher annual interest rates decrease the likelihood of success. Previous studies show that elevated interest rates correlate with higher loan default rates, impacting investor decisions \citep{polena_determinants_2018}. 
In addition, borrower characteristics play vital roles; older individuals and those with a credit history tend to have higher success rates. 
Financial indicators, such as income level and car ownership, also significantly affect fundraising outcomes. For comparison, we apply the Oracle method, which confirms these findings, underscoring the effectiveness of the proposed method.

\begin{table}
\TABLE
{Analysis of financial platform data: the variable selection and estimation results for the proposed estimation. \label{tab:real_data_estimation}}
{
\begin{tabular}{p{100pt}p{150pt}ccc}  
\toprule 
Variable                 & Description                                                                    & Subgroup 1 & Subgroup 2 & Subgroup 3  \\
\midrule 
BidAmount                & The amount of money for the fundraising bid                                    & -0.348 & -0.485 & -0.620 \\
APR                      & Annual interest rate                                                           & -0.287 & -0.338 & -0.330 \\
Age                      & Age                                                                            & 0.310  & 0.307  & 0.574  \\
LoanNumbers              & The number of successful loans                                                 & 0.278  & 0.358  & 0.435  \\
RepaymentPeriod\_2 & Repayment period for the bid (6 to 12 months)                     & -0.331 & -0.472 \\
RepaymentPeriod\_3 & Repayment period for the bid (12 to 18 months)                                 & -0.885 &        &        \\
sex\_2                   & Sex: female                                                                    & -0.252 & -0.771 & -0.647 \\
MaritalStatus\_1         & Marital status: spinsterhood                                                   & -0.229 &        & -0.156 \\
SuccessfulNum\_1         & The number of previous successful fundraising bids (equals 1)                  & 2.157  & 1.773  & 2.173  \\
SuccessfulNum\_2         & The number of previous successful fundraising bids (equals 2)                  & 1.212  & 1.201  &        \\
SuccessfulNum\_6         & The number of previous successful fundraising bids (larget than or equals 6) & 0.479  &        & 0.886  \\
IncomeLevel\_1           & Monthly Income level (Below 2000   yuan)                            & -0.106 &        \\
IncomeLevel\_3           & Monthly Income level (Below 2002 yuan)                                         & 0.204  &        &        \\
hasCar\_2                & The car ownership of the borrower                                                              & 0.115  &        &        \\
jobType\_1               & Job type of borrower (salariat)                                                & -0.423 & -0.265 & -0.468
 \\
\bottomrule  
\end{tabular}
}{}
\end{table}

\begin{table}
\TABLE
{Analysis of financial platform data: test AUC results for different provinces.\label{tab:real_data_auc}}
{
\begin{tabular}{r*{5}{p{0.16\textwidth}<{\centering}}}
\toprule 
  & Oracle & Ind & Homo &  Proposed \\
  \midrule  
 Guangdong & 0.966 & 0.954 & 0.932 & 0.961 \\
Zhejiang  & 0.949 & 0.740 & 0.931 & 0.934 \\
Shandong  & 0.962 & 0.850 & 0.939 & 0.954 \\
Fujian    & 0.949 & 0.900 & 0.917 & 0.946 \\
Hunan     & 0.933 & 0.890 & 0.878 & 0.943 \\
Hubei     & 0.968 & 0.601 & 0.904 & 0.944 \\
Sichuan   & 0.935 & 0.590 & 0.927 & 0.942 \\
Henan     & 0.963 & 0.887 & 0.916 & 0.962 \\
Beijing   & 0.932 & 0.911 & 0.906 & 0.909 \\
Hebei     & 0.962 & 0.939 & 0.930 & 0.965 \\
Anhui     & 0.963 & 0.661 & 0.931 & 0.962 \\
Guangxi   & 0.948 & 0.798 & 0.934 & 0.934 \\
Jiangxi   & 0.940 & 0.747 & 0.929 & 0.952 \\
\midrule  
Average   & 0.952 & 0.805 & 0.921 & 0.947 \\
\bottomrule
\end{tabular}
}
{}
\end{table}

To evaluate the predictive performance of the proposed method, we split the data into training and testing sets, using batches 1 to 10 serve for training and batches 11 to 13 for testing. Table \ref{tab:real_data_auc} reports the test AUC values for various models across different provinces, along with their averages.
The average test AUC for the proposed method is 0.947, compared to 0.952 for Oracle, 0.805 for Ind, and 0.921 for Homo. 
This suggests that the proposed method outperforms the alternatives and is comparable to the infeasible Oracle method. 
This trend is consistent across provinces, with the proposed method generally achieving higher AUC values than the alternatives. Additionally, Figure S7 of the supplementary material presents the corresponding trace plot, indicating that the proposed method consistently surpasses the Ind method and progressively approaches the performance of the Oracle method. In summary, these findings highlight the predictive capabilities of the proposed method in real-world applications.  

\subsection{Analysis of web log data}
We apply the proposed method to analyze bank web log data to detect potential cyberattacks by identifying abnormal requests directed at its portal websites. 
In general, users interact with websites mainly through Hypertext Transfer Protocol (HTTP) requests, most of which are ``normal" from legitimate users, while some are ``abnormal", indicating possible cyberattacks. 
The web logs contain parameters submitted as a series of key-value pairs, with sample logs available in Table S11 of the supplementary material. 
Traditional security products, like intrusion prevention systems (IPS) and web application firewalls (WAF), primarily use comparison-based approaches, which can struggle with unknown or sophisticated attacks, such as zero-day exploits. In contrast, our method extracts features from web logs to develop statistical models for more flexible, real-time cyberattack detection.

The dataset contains parts of the web logs from $K=11$ websites (named as \textit{url1}--\textit{url11}), collected over 16 days (August 29 to September 9, 2019), with varying log volumes. 
The total samples for the websites range from 1,150 to 3,368, and we analyze a binary response variable where $y_{i}=1$ indicates abnormal requests and $y_{i}=0$ indicates normal requests. The overall abnormal rate is 34.85\%, with individual rates between 24.40\% and 50.06\%. 
We performed feature engineering to generate $p=45$ numerical predictors from the web logs, including total request length (Plen), parameter (key-value pairs) counts (Pnum), the lengths of various parameters (Pl0--Pl19), and the descriptions about various parameter types (Pcnt\_num, Pratio\_num, Pt0\_str, Pt0\_num, and Pt1\_not\_na--Pt19\_not\_na).

Using the proposed method, we group the 11 websites into 5 subgroups of sizes 4, 3, 2, 1 and 1, with models involving 9, 6, 28, 6 and 7 variables, respectively, showing substaintial overlaps. The detailed results of the variable selection and estimation are summarized in Table S13 of the supplementary material. 
In particular, variables Pl0-Pl2 (lengths of the first to fourth parameters in the request) are selected for more than 4 subgroups with positive signs, indicating that abnormal requests may contain redundant contents. 
Other frequently selected variables include Pcnt\_num (the number of parameters with type ``num"), Pt0\_str (whether the type for the first parameter is ``str"), Pt0\_not\_na (whether the type for the first parameter is not ``na"), and others. These findings provide valuable insights on cyberattack patterns.  

To evaluate performance, batches 1 to 12 serve as the training dataset, while batches 13 to 15 are for testing.
The results show that the average test AUC for the proposed method is 0.866, compared to 0.872 for Oracle, 0.857 for Ind, and 0.851 for Homo, demonstrating that the proposed method outperforms other alternatives and is comparable to the Oracle method. 
Detailed test AUC results for different websites and trace plots for the average test AUC across batches are provided in Table S14 and Figure S8 of the supplementary material.
Overall, these results further confirm the effectiveness of the proposed method in practical applications.

\section{Discussion and Conclusions} \label{sec:discussion}
This article presents a federated online method for high-dimensional heterogeneous multisource streaming data analysis. We address issues of homogeneity and heterogeneity among data sources using a subgroup assumption for modeling. 
Our proposed penalized renewable estimation and proximal gradient descent algorithm simultaneously estimates coefficients, identifies important variables, and recovers subgroup structure among data sources. It does not require raw data transmission across sources and only relies on the current batch raw data and summary statistics from previous batches for model updates. This intuitive approach has several well-desired theoretical properties and offers a unique modeling strategy compared to existing approaches.

Future research directions include incorporating prior information, such as network connections or tree structures between data sources, to enhance multisource modeling. Additionally, we currently assume synchronous data arrival; developing approaches for asynchronous scenarios, where updates occur at different times, poses further challenges. Finally, while our methodology focuses on Generalized Linear Models (GLMs), it could be adapted for other statistical models like the Cox model or finite-mixture models.



\bibliographystyle{informs2014} 
\bibliography{my_ref} 





\end{document}